\documentclass[conference]{IEEEtran}
\IEEEoverridecommandlockouts
\usepackage[moderate, tracking = normal]{savetrees}
\usepackage{cite}
\usepackage{amsmath,amssymb,amsfonts}
\usepackage{algorithmic}
\usepackage{graphicx}
\usepackage{epstopdf}
\usepackage{url}
\usepackage{svg}
\usepackage{textcomp}
\usepackage{xcolor}
\usepackage{natbib}
\usepackage{xspace}

\newcommand{\digit}{\protect{$\mathcal{D}ig{IT}$}\xspace}
\def\BibTeX{{\rm B\kern-.05em{\sc i\kern-.025em b}\kern-.08em
    T\kern-.1667em\lower.7ex\hbox{E}\kern-.125emX}}
\begin{document}

\title{Architecting Digital Twins for Intelligent Transportation Systems\\
}





\author{
\IEEEauthorblockN{Hiya Bhatt\IEEEauthorrefmark{1}, Sahil\IEEEauthorrefmark{2}, Karthik Vaidhyanathan\IEEEauthorrefmark{1}, Rahul Biju\IEEEauthorrefmark{2}, Deepak Gangadharan\IEEEauthorrefmark{2}, Ramona Trestian\IEEEauthorrefmark{3}, Purav Shah\IEEEauthorrefmark{3}} 
\IEEEauthorblockA{\IEEEauthorrefmark{1}\textit{Software Engineering Research Centre, IIIT Hyderabad, India} \{hiya.bhatt@research.iiit.ac.in, karthik.vaidhyanathan@iiit.ac.in\}}
\IEEEauthorblockA{\IEEEauthorrefmark{2}\textit{Computer Systems Group, IIIT Hyderabad, India} \{sahil.sahil@research.iiit.ac.in, rahul.biju@research.iiit.ac.in, deepak.g@iiit.ac.in\}}
\IEEEauthorblockA{\IEEEauthorrefmark{3}\textit{Faculty of Science and Technology, Middlesex University London} \{r.trestian@mdx.ac.uk, p.shah@mdx.ac.uk\}}
}

\maketitle
\begin{abstract}
Modern transportation systems face growing challenges in managing traffic flow, ensuring safety, and maintaining operational efficiency amid dynamic traffic patterns. Addressing these challenges requires intelligent solutions capable of real-time monitoring, predictive analytics, and adaptive control. This paper proposes an architecture for DigIT, a Digital Twin (DT) platform for Intelligent Transportation Systems (ITS), designed to overcome the limitations of existing frameworks by offering a modular and scalable solution for traffic management. Built on a \textbf{Domain Concept Model (DCM)}, the architecture systematically models key ITS components enabling seamless integration of predictive modeling and simulations. The architecture leverages machine learning models to forecast traffic patterns based on historical and real-time data. To adapt to evolving traffic patterns, the architecture incorporates adaptive Machine Learning Operations (MLOps), automating the deployment and lifecycle management of predictive models. Evaluation results highlight the effectiveness of the architecture in delivering accurate predictions and computational efficiency.


\end{abstract}

\section{Introduction}  

The increasing complexity of modern transportation systems, characterized by growing vehicular density, diverse communication protocols, dynamic and distinct traffic patterns, and complex road networks poses significant challenges in traffic management, safety, and operational efficiency. To this end, Intelligent Transport System (ITS) has emerged as a potential solution as they serve as an ecosystem that integrates technology, communication, and data analytics to improve the efficiency, safety, and sustainability of transportation systems.
ITS solutions are used to manage and optimize traffic flow, enhance user experience, and support advanced applications like autonomous driving and vehicular communication~\cite{perallos2015intelligent, andersen2000intelligent}. Most importantly, the ITS infrastructure helps reduce the accidents that occur on the roads. Hence, it is important to deploy more ITS infrastructure as fatal accidents are on the rise in several countries~\cite{NHTSA_Traffic_Stats, Road_Accidents_India}.

However, deploying an ITS infrastructure, which achieves the earlier mentioned objectives while ensuring safety, is non trivial due to its scale, dynamic environment, and real-time demands. The development of a formally sound ITS solution thus becomes challenging as it is difficult to verify all the what-if scenarios in the physical deployment. In response, Digital Twins (DTs) have emerged as a transformative paradigm, enabling the creation of virtual replicas of physical systems to support real-time monitoring, predictive analytics, and adaptive control mechanisms \cite{grieves2017digital}\cite{liu2021review}\cite{barricelli2019survey}\cite{feng2023resilience}. In the context of ITS, DTs leverage real-time data from IoT-enabled sensors, vehicular networks, and environmental monitoring systems to improve situational awareness and enhance decision-making processes \cite{jafari2023review}. Existing implementations, such as Mobility DT \cite{mobilityWang}, have demonstrated their utility in modeling traffic flows, evaluating rerouting strategies, and predicting congestion patterns through simulation-based approaches. These advancements highlight the potential of DTs to transform urban mobility and transportation planning.  

Although modeling traffic flows works well in several countries, the challenge is significantly higher in many other countries that witness highly unstructured traffic. This scenario also demands robust traffic management solutions to mitigate the social, economic, and environmental impacts of congestion. Traffic flow prediction enables proactive decision-making and efficient traffic control. However, the irregular and dynamic nature of traffic patterns, presents unique challenges in identifying consistent trends. To address these complexities, this project leverages temporal and spatio-temporal Deep Learning (DL) models, such as LSTM (Long Short-Term Memory) \cite{lstmbase} and BiLSTM (Bidirectional LSTM) \cite{bilstm}. LSTM, a variant of Recurrent Neural Networks (RNNs), excels in capturing long-term dependencies in sequential data, making it highly effective for modeling traffic patterns that evolve over time. Its ability to retain and utilize information over long sequences enables accurate predictions of future traffic flows based on historical trends. By integrating real-time data from traffic nodes at intersections in Hyderabad, India, the system ensures precise predictions and real-time adaptability. Automated retraining mechanisms further enhance robustness, contributing to smarter and more efficient traffic management systems.

This paper introduces architecture for \digit (DT for ITS) platform, addressing the limitations of existing frameworks by providing a modular, scalable, and adaptive solution for traffic management. To systematically model the components and behaviors of ITS, the architecture adopts a \textit{Domain Concept Model (DCM)}, which captures key entities such as vehicles, sensors, communication networks, and user behaviors. The DCM serves as the foundation for designing the architecture and defining the interactions required to simulate traffic scenarios and predict outcomes effectively.  

The proposed architecture integrates real-time data streams with Machine Learning (ML) based predictive modeling to optimize traffic flow and enhance decision-making. It employs ML models to forecast traffic patterns based on historical and real-time data. The predictions are validated through simulations, enabling scenario testing and evaluation of intervention strategies. By bridging modeling and simulation, this architecture demonstrates the potential of integrating predictive analytics into real-time traffic management workflows. The architecture for \digit platform incorporates adaptive \textit{Machine Learning Operations (MLOps)} \cite{symeonidis} practices, automating the deployment and lifecycle management of predictive models. This framework supports dynamic adjustments to evolving traffic conditions while maintaining computational efficiency and scalability. The results presented in this work validate the system’s accuracy and responsiveness, demonstrating its suitability for real-world ITS applications.

\section{Related Work}

Digital Twin (DT) technology has shown significant potential across diverse domains by enabling real-time monitoring, predictive analytics, and scenario-based simulations. Its adaptability has made it particularly effective in applications ranging from infrastructure management to smart cities, offering a foundation for addressing the challenges of Intelligent Transportation Systems (ITS).

In infrastructure management, Likhit et al. \cite{kanigolla2024architecting} demonstrated DTs' effectiveness in optimizing water distribution networks through real-time data and simulations. Their work highlights leveraging predictable flow patterns to improve resource management. However, transportation systems add complexity due to unpredictable traffic dynamics and human behavior, as noted by Glaessgen and Stargel \cite{glaessgen2012digital}, requiring robust frameworks to address uncertainties.

Expanding to urban-scale applications, Mohammadi et al. \cite{MohammadiSmartCity} applied DTs for smart city optimization, enabling better decision-making and resource allocation. Similarly, Barat et al. \cite{barat2022agent} modeled pandemic dynamics, emphasizing adaptability in crisis scenarios. Despite these advances, Barn et al. \cite{balbir} highlighted gaps in integrating socio-technical interactions and behavioral modeling, limiting realism in existing frameworks.

To address scalability and interoperability, Redelinghuys et al. \cite{redelinghuys2020six} proposed layered DT architectures, promoting modularity. Complementing this, Kulkarni et al. \cite{kulkarni2023dtworkflow} introduced automated workflows for predictive modeling, ensuring responsiveness under evolving conditions. Yet, these methods often rely on static configurations, posing challenges for dynamic systems requiring continuous retraining.

Despite advancements, Digital Twin (DT) systems still struggle with real-time synchronization between virtual and physical environments, often relying on offline simulations \cite{baolixia}. Recent efforts, such as Ge et al. \cite{kuvsic2023digital}, introduced cyber-physical frameworks with IoT sensors for synchronized traffic simulations. However, scalability to multi-modal urban systems remains a challenge \cite{lee2015cyber}.

To tackle these limitations, Domain-Driven Design (DDD) offers a modular approach for structuring complex DTs. By leveraging bounded contexts, DDD improves scalability and adaptability. Macias et al. \cite{macias2023architecting} applied DDD principles for scalable architectures, Evans \cite{evans2004domain} emphasized adaptability and automated updates, highlighting DDD's potential to meet real-time demands and evolving requirements.

Building on this foundation, we propose an architecture for the \digit platform—a DT framework for ITS that integrates predictive modeling, simulations, and automated workflows. The architecture applies DDD principles to identify domains, which are modeled using a Decision-Component Model (DCM). The DCM captures key entities such as vehicles, sensors, communication networks, and user behaviors, ensuring seamless integration between physical and digital systems. ML models handle traffic forecasting, supported by MLOps pipelines for scalability through automated retraining. Simulations enable scenario testing, providing insights for decision-making and interventions. By combining these elements, our framework addresses modern transportation complexities with a dynamic and scalable approach.

\section{ITS Deployment Case Study}
\label{sec:usecase}
The creation of a Digital Twin (DT) for Intelligent Transportation Systems (ITS) entails the deployment of IoT-enabled traffic monitoring infrastructures in urban settings. This approach aims to establish a robust and scalable framework for analyzing and enhancing traffic flow efficiency. The primary objectives include real-time data acquisition, traffic scenario simulation, and predictive modeling to tackle issues such as congestion, delays, and inefficiencies in traffic control.

For illustrative purposes, an ITS prototype has been implemented through a network of edge devices positioned strategically at high-traffic intersections near the IIIT Hyderabad campus. These devices are outfitted with cameras and processing units that gather real-time traffic metrics, such as vehicle counts, density levels, and congestion indicators, at regular time intervals. Advanced temporal models, including Long Short-Term Memory (LSTM) and Bidirectional LSTM (BiLSTM), are utilized for on-device data processing. These models excel in recognizing temporal patterns in traffic data and forecasting future trends, leveraging LSTM's capability to capture long-term dependencies and BiLSTM's strength in analyzing data sequences from both directions.

To enhance data quality and storage efficiency, the processed information is consolidated into 5-minute intervals before being transmitted to a public API hosted on a Virtual Private Server (VPS). This centralized API facilitates seamless data access for various system components. Moreover, the localized processing on edge devices reduces latency, enabling prompt responses to dynamic traffic situations.

This configuration offers continuous, real-time insights into traffic dynamics, particularly valuable during peak hours or unforeseen events like accidents or road closures. The combination of LSTM and BiLSTM models significantly improves the system's adaptability to fluctuating traffic conditions. The architecture also supports periodic retraining of the models, triggered either by scheduled routines or performance-based criteria, ensuring sustained accuracy and relevance of predictions.

The accumulated data serves as the foundation for simulations conducted with tools like the Simulation of Urban MObility (SUMO) \cite{SUMO2018}, which replicate actual traffic scenarios to evaluate vehicle behavior at monitored intersections. These simulations can assess the impact of strategies such as modifying traffic light schedules or redirecting traffic flow to alleviate congestion. Additionally, machine learning models analyze both real-time and historical datasets to forecast future traffic conditions, identifying trends like anticipated congestion spikes during evening rush hours.

This case study illustrates the potential of DT in enhancing traffic management in urban environments. By integrating real-time data monitoring, scenario simulation, and predictive analytics, the system equips traffic authorities with the capability to observe current conditions and virtually test potential interventions before applying them in the real world. Throughout this paper, we refer to this case study to contextualize and elaborate on the architecture proposed for developing and deploying a DT in ITS applications.

\section{Architecture for Digital Twin}
\label{sec:architecture}
The design and implementation of the architecture of \digit (DT for ITS) platform require a structured approach to address the inherent complexities of urban mobility. Using the Sociotechnical DT Specification Meta Language Concepts \cite{balbir} as the foundation, this approach establishes the scope, purpose, and operational mechanisms of the DT. At the core of this system is the Domain Concept Model (DCM), which provides a structured representation of the architecture of \digit components and their interactions. Together, these elements ensure that the architecture for \digit platform is both theoretically sound and practically viable. 

\subsection{Sociotechnical DT Specification Meta Language Concepts for DigIT}

The specification meta-language serves as a framework to articulate and guide the development of the DT for ITS. Each concept within this framework is defined and applied explicitly to address the challenges and requirements of ITS.

\begin{figure*}[!htbp]
    \centering
    \includegraphics[width=\linewidth]{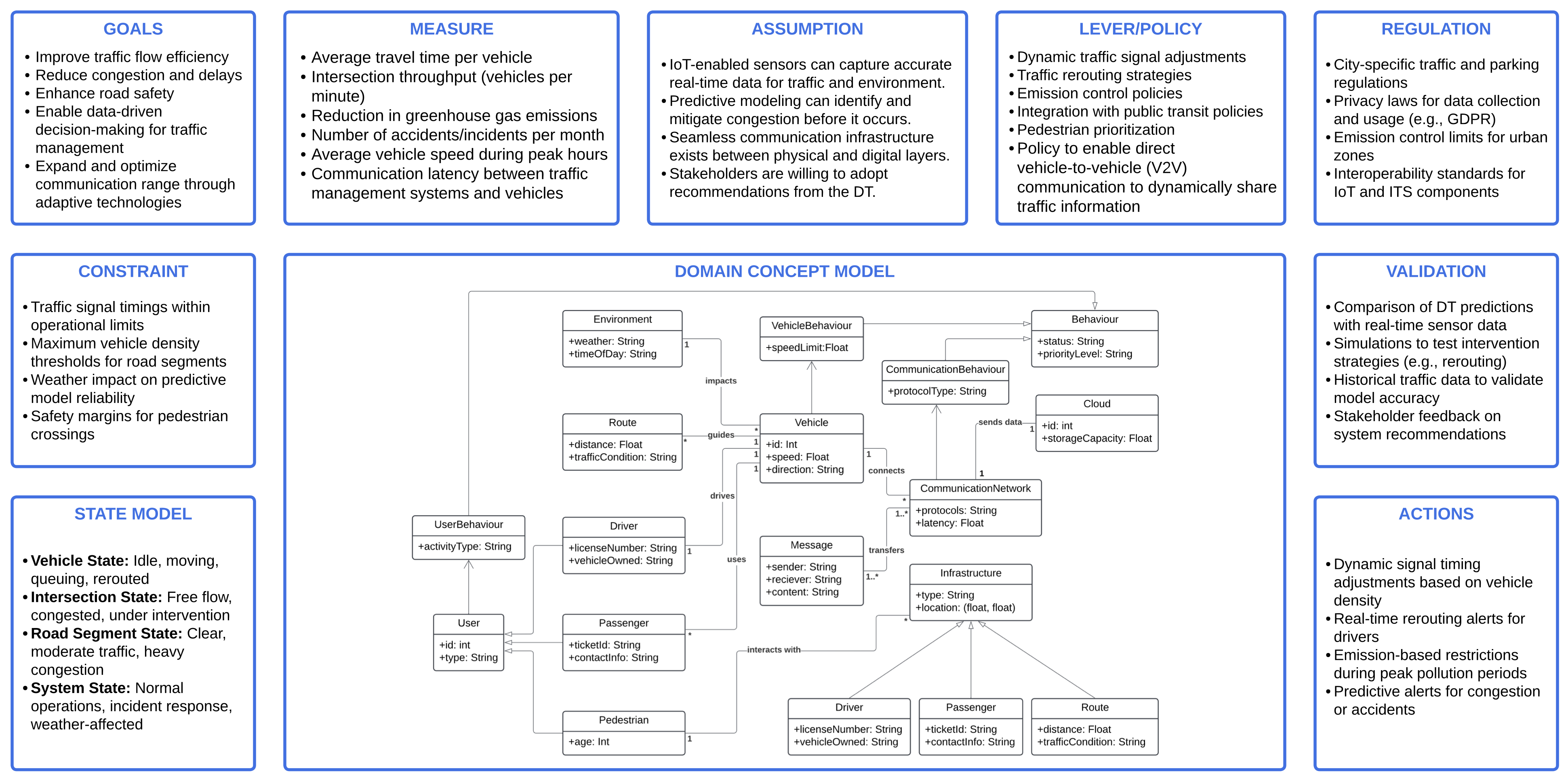}
    \caption{Sociotechnical DT Specification Meta Language Concepts for ITS.}
    \label{fig:dcm_uml}
\end{figure*}

\textbf{Goal.} The primary goal of the DT is to enhance traffic management by enabling proactive and data-driven decision-making. Key objectives include improving traffic flow efficiency, reducing congestion and delays, and enhancing road safety. Additionally, the DT seeks to expand and optimize communication capabilities through adaptive technologies, ensuring seamless information exchange between vehicles and infrastructure. These goals collectively aim to create a more responsive and sustainable ITS. 

\textbf{Measure.} The effectiveness of the DT is evaluated using quantifiable metrics that assess traffic performance, environmental impact, and communication reliability. These measures include average travel time per vehicle, intersection throughput (vehicles per minute), reduction in greenhouse gas emissions, and the number of accidents or incidents reported per month. Additional metrics, such as average vehicle speed during peak hours and communication latency between traffic management systems and vehicles, provide insights into system responsiveness and efficiency. These indicators ensure the DT delivers measurable improvements aligned with ITS objectives.

\textbf{Assumption.} The assumptions explain the scope of the DT and define the prerequisites for its successful deployment. For the ITS-focused DT, the architecture assumes the presence of IoT-enabled sensors deployed across the transportation network. These sensors capture real-time data, such as traffic flow rates, vehicular speeds, and congestion levels. Environmental sensors are also assumed to provide supplementary data on external factors like weather conditions. Additionally, the architecture assumes the existence of a robust communication infrastructure that facilitates the seamless transmission of data between the physical and digital layers. These assumptions ensure that the DT operates in a data-rich environment, accurately reflecting real-world conditions \cite{9745481}.

\textbf{Lever/Policy.} Levers or policies are quantifiable or actionable elements that influence the operation of the DT. Examples of policies include dynamic traffic signal adjustments, traffic rerouting strategies, emission control policies, integration with public transit policies, pedestrian prioritization, and enabling direct vehicle-to-vehicle (V2V) communication for dynamic traffic information sharing. These levers serve as input for simulations and predictive models, allowing the DT to anticipate traffic conditions and recommend appropriate interventions. For example, a high vehicle density at a particular intersection could be detected and processed by an ML model, which predicts the potential impact of congestion. This prediction could then initiate simulations to assess the effectiveness of rerouting strategies or signal adjustments to alleviate the congestion.

\textbf{Regulation.} Regulations define enforceable policies and standards that guide the operations of the DT within the ITS. These include city-specific traffic and parking rules, privacy laws governing data collection and use, emission control limits for urban zones, and interoperability standards for IoT and ITS components. Such regulations ensure that the DT complies with societal, environmental, and legal requirements. For example, emission control policies can trigger restrictions on vehicle performance during periods of high pollution, while interoperability standards ensure seamless integration between heterogeneous systems within the ITS ecosystem.

\textbf{Constraint.} Constraints define domain-specific rules that govern the behavior of the DT to ensure compliance with legal, ethical, and operational standards. In the context of ITS, these constraints include traffic signal timings that must remain within operational limits, maximum vehicle density thresholds for road segments, and safety margins for pedestrian crossings. Additionally, environmental factors, such as weather conditions, impose constraints on the reliability of predictive models. By embedding these constraints into its architecture, the DT operates within predefined boundaries, ensuring safe, reliable, and regulation-compliant functionality.

\textbf{Validation.} The DT is validated using a multifaceted approach to ensure accuracy and reliability. Real-time sensor data collected from IoT-enabled devices at intersections are compared against DT predictions to assess model performance. Simulations are conducted to test intervention strategies, such as rerouting, under varying traffic conditions. In addition, stakeholder feedback on system recommendations provides qualitative insights, ensuring that the DT meets practical requirements and operational expectations.


\textbf{Actions.} Actions represent the operational behaviors assigned to domain concepts within the DT. These include dynamic signal timing adjustments, traffic rerouting based on real-time conditions, and generating predictive alerts for traffic operators. By translating the DT's insights into actionable interventions, these behaviors create a feedback loop that continuously refines the system's performance.

\textbf{State Model.} The State Model defines the dynamic conditions of key entities within the ITS to capture their operational status and transitions. Vehicles can be in states such as idle, moving, queuing, or rerouted. Intersections are classified as free flow, congested, or under intervention. Road segments are categorized based on traffic levels, ranging from clear to heavy congestion. The overall system state reflects conditions like normal operations, incident response, or weather-affected scenarios. These states enable real-time monitoring and adaptive decision-making within the DT framework.

\textbf{Domain Concept Model (DCM).} DCM defines the ITS as a set of interconnected components that capture attributes, behaviors, and relationships at a higher level of abstraction. It provides a foundational framework for modeling key entities such as vehicles, users, communication networks, infrastructure, routes, and environmental conditions, along with their interactions. Each component in the DCM can be further elaborated through specialized domain models to address specific aspects of ITS operations in greater detail. We used the principles of \textit{Domain-Driven Design (DDD)} to identify key domains in the ITS. This was achieved by identifying \textit{bounded contexts} within the ITS domain, as proposed by Macías et al.~\cite{macias2023architecting} and Evans ~\cite{evans2004domain}. These bounded contexts provide logical partitions that encapsulate specific functionalities, ensuring modularity and consistency across the system.

Building on this foundation, the model organizes the ITS into several interconnected components, each corresponding to a bounded context as shown in Figure~\ref{fig:dcm_uml}. For instance, the \textit{Users Context} categorizes users as drivers, passengers, and pedestrians, each with distinct attributes and behaviors. Similarly, the \textit{Vehicles Context} models properties such as speed, direction, and unique identifiers, as well as behaviors like adherence to speed limits, to capture real-world dynamics. The \textit{Communication Context} defines networks and protocols to enable seamless data exchange between vehicles, infrastructure, and centralized systems, supporting real-time information sharing and decision-making. Environmental factors, such as weather and time of day, influence traffic conditions and vehicle performance, while infrastructure components define fixed resources like traffic signals and IoT-enabled sensors. Messages exchanged through communication protocols support coordination, while routes and traffic conditions guide vehicle navigation and behavior. The DCM also models interactions and constraints, including traffic signal timing limits, vehicle density thresholds, and pedestrian safety margins. These constraints are embedded within the architecture to ensure compliance with regulatory and operational requirements. Behavioral elements, such as adaptive signal timing, dynamic rerouting, and message exchanges, enable predictive analytics and scenario testing within the DT. By abstracting ITS components at a high level, DCM allows modular expansion, enabling detailed exploration of specific domains through separate models. Figure~\ref{fig:dcm_uml} illustrates the structure, showing how core entities and relationships form the basis for an extensible and scalable ITS framework. It is important to note that this DCM is not exhaustive but rather serves as an illustrative example of how the ITS domain can be modeled. The framework is flexible and can be extended or customized to incorporate additional components and interactions as needed to address the specific requirements of diverse ITS applications.


\subsection{Proposed Architecture of the DT}  
Building on the DCM, we propose an architecture for \digit platform that transforms the high-level abstractions defined in the DCM into operational components for data collection, predictive analysis, simulation and visualization. Although DCM provides a broad framework for modeling ITS, this architecture focuses on select components essential for real-time traffic monitoring, prediction, and scenario-based simulations. Using elements from the DCM, such as vehicles, communication networks, sensors, and user behaviors, the proposed architecture ensures a seamless integration between conceptual modeling and practical implementation. The following sections detail each functional component, demonstrating how the principles outlined in the DCM are operationalized within the architecture of \digit platform.

\begin{figure}[!htbp]
    \centering
    \includegraphics[width=0.90\linewidth]{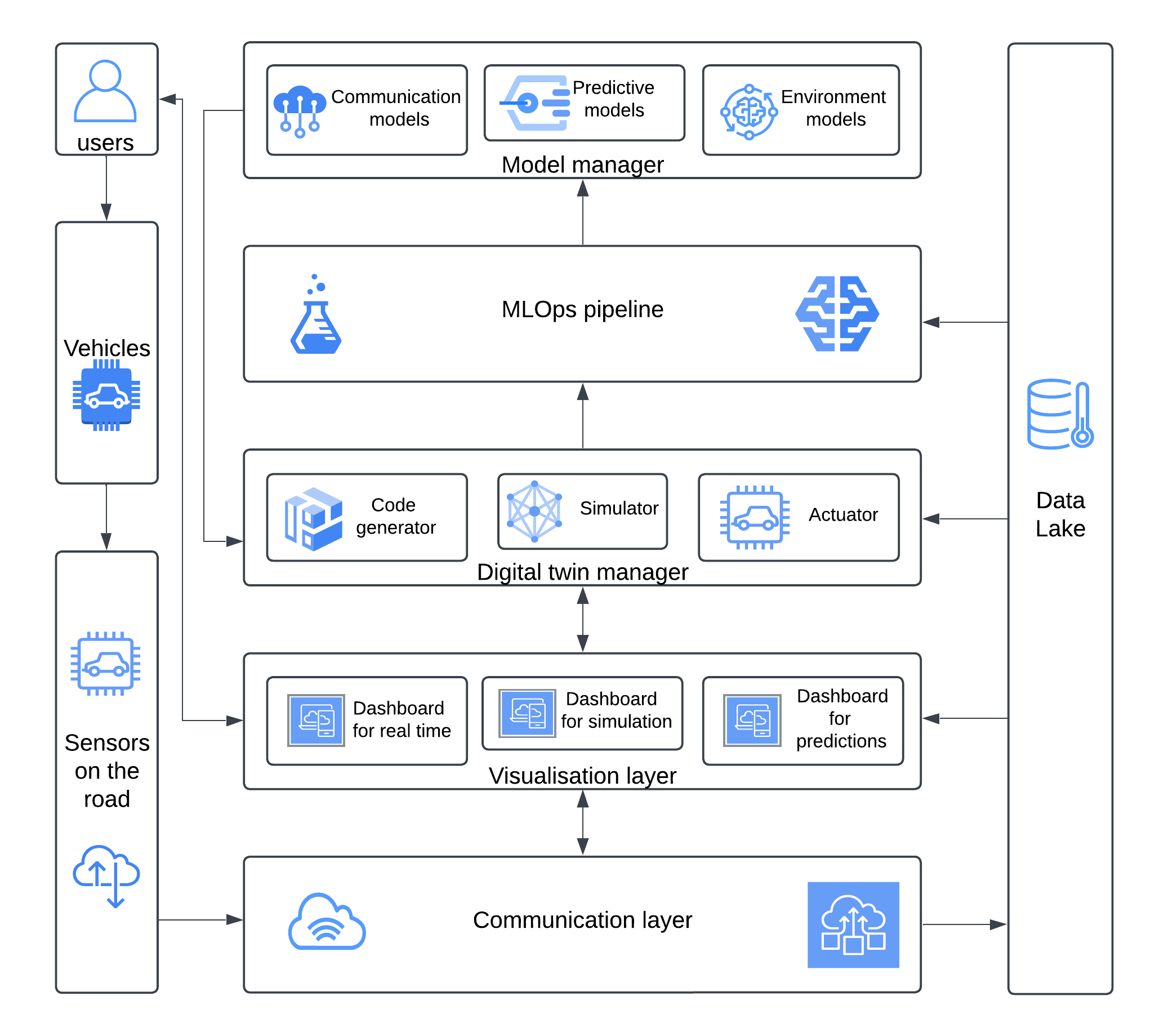}
    \caption{Architecture of \digit}
    \label{fig:digit_architecture}
\end{figure}

\textbf{Users and Vehicles.}  
As shown in Figure~\ref{fig:digit_architecture}, the architecture models users, including drivers, passengers, and pedestrians, along with vehicles characterized by attributes such as speed, direction, and ID. Users access dashboards that provide real-time traffic updates and predictions, extending the behavioral modeling of users and vehicles defined in the DCM to ensure synchronization between physical and digital systems.

\textbf{Sensors.}  
As shown in Figure~\ref{fig:digit_architecture}, IoT-enabled sensors deployed at intersections and road segments capture real-time data, including vehicle speeds, traffic density, and congestion levels. Environmental sensors complement this by monitoring external factors such as weather conditions, aligning with the \textit{Environment} component in the DCM (Figure~\ref{fig:dcm_uml}). These sensors operate continuously, reflecting the dynamic nature of the transportation network and providing input for predictive modeling and simulations. By integrating data from physical infrastructure, the sensors ensure that the DT maintains synchronization with real-world conditions, enabling timely analysis and decision-making.

\textbf{Communication Layer.}  
The communication layer, as shown in Figure~\ref{fig:digit_architecture}, facilitates seamless and reliable data exchange between the physical and digital systems. It leverages a heterogeneous network comprising 4G/5G, LoRA, and WiFi communication mechanisms to support high-frequency real-time data streams and periodic batch transfers. Modeled in the DCM (Figure~\ref{fig:dcm_uml}) through the \textit{CommunicationNetwork} and \textit{CommunicationBehaviour} components, this layer ensures bidirectional communication, enabling the DT to relay recommendations—such as optimized signal timings or rerouting strategies—back to the physical infrastructure. Attributes such as protocol type and latency, as captured in the DCM, align with the low-latency protocols implemented here to handle time-critical updates, including responses to accidents or congestion. For example, vehicle counts and speed data collected at multiple intersections are transmitted to the data lake while maintaining synchronization across all input sources.

\textbf{Data Lake.}  
After data is transmitted through the communication layer, the data lake stores the communicated information, supporting both real-time analysis and long-term storage. It processes structured and unstructured data, performing preprocessing tasks like cleaning and aggregation to optimize it for simulations and machine learning models. As shown in Figure~\ref{fig:digit_architecture}, this layer integrates inputs modeled in the DCM, including vehicle behavior, environmental conditions, and communication protocols.

\textbf{Visualization.}  
As shown in Figure~\ref{fig:digit_architecture}, the visualization layer provides three types of dashboards—\textit{real-time, simulation,} and \textit{prediction}. The real-time dashboard monitors live metrics like vehicle counts and congestion levels. The simulation dashboard models scenarios such as accidents and peak-hour traffic, while the prediction dashboard forecasts future traffic patterns using historical and real-time data. These dashboards leverage the behavioral and environmental models defined in the DCM to ensure synchronization between real-world data and simulated outputs.

\textbf{Digital Twin Manager.}  
As shown in Figure~\ref{fig:digit_architecture}, the DT Manager creates a virtual representation of the physical transportation network. It enables the simulation, prediction, and optimization of traffic conditions in real-time, while ensuring continuous feedback between the virtual and physical systems. The DT Manager incorporates three key components: the \textit{Simulator}, the \textit{Actuator}, and the \textit{Code Generator}. These components collaboratively ensure the platform's adaptability, scalability, and responsiveness to evolving traffic conditions.

The \textit{Simulator} replicates real-world traffic conditions by using real-time and historical data from IoT-enabled sensors, as described in the Figure~\ref{fig:digit_architecture}. It models traffic dynamics such as vehicle movements, congestion patterns, and disruptions (e.g., accidents or roadblocks). Scenarios are evaluated using predictive models, such as LSTM and BiLSTM, to assess the impact of potential interventions like signal timing adjustments, lane closures, or rerouting strategies. For instance, when a sensor detects congestion at an intersection, the Simulator evaluates alternative traffic flow patterns and identifies the most effective mitigation strategy. The simulations are run iteratively, incorporating temporal and spatial dependencies, ensuring the virtual environment remains in sync with the physical system.

The \textit{Actuator} implements decisions derived from the Simulator into the physical transportation network. By translating simulation outcomes into real-world actions, it controls physical infrastructure such as traffic lights, dynamic message signs, and variable speed limits. For example, if the Simulator predicts bottlenecks at a specific intersection, the Actuator adjusts traffic signal timings or reroutes vehicles to alleviate congestion. The Actuator is designed to handle low-latency operations, ensuring rapid deployment of interventions during critical scenarios, such as accidents or emergency evacuations. The bidirectional communication with physical systems ensures that the Actuator can validate its actions using updated feedback from sensors, maintaining a robust closed-loop system.

The \textit{Code Generator} serves as the intermediary between the high-level abstract models defined in the DCM and the operational components of the Digital Twin. While ML models can directly process data for predictions without translation, abstract models in the DCM—such as communication protocols, behavioral models, and state transitions—must be translated into executable instructions for the Digital Twin to understand and act upon them. The Code Generator performs this crucial task, converting these specifications into machine-readable configurations that the Simulator and Actuator can execute. For example, the DCM may define the relationships between vehicles, sensors, and infrastructure; the Code Generator ensures these relationships are accurately mapped to system operations, enabling real-time simulation and execution. This modular approach ensures flexibility and scalability, allowing for the seamless integration of new traffic management policies or sensor types without disrupting the existing system.
\textbf{MLOps Pipeline.}  
The MLOps pipeline automates the lifecycle of machine learning (ML) models, such as Long Short-Term Memory (LSTM) for traffic flow prediction or Convolutional Neural Networks (CNN) for anomaly detection. It handles data preprocessing, model training, deployment, and monitoring, ensuring adaptability to evolving traffic conditions. Let $X_t = \{x_1, x_2, \dots, x_t\}$ represent input data collected over $t$ time intervals. The objective is to forecast traffic states for a future horizon $h$ as:
\vspace{0.5mm}
\[
\hat{Y}_{t+h} = f(X_t; M_v)
\]

\noindent where $M_v$ denotes the current model version. Model performance is continuously evaluated using metrics such as Root Mean Squared Error (RMSE):
\vspace{0.5mm}
\[
RMSE = \sqrt{\frac{1}{n} \sum_{i=1}^{n} (Y_i - \hat{Y}_i)^2}.
\]

\noindent To maintain model accuracy, retraining is triggered based on drift detection. Drift $\delta$ is defined as:
\vspace{0.5mm}
\[
\delta = D(Y, \hat{Y}) > \epsilon,
\]

\noindent where $D(\cdot)$ measures data or model drift, such as distributional changes in input features or output predictions, and $\epsilon$ represents a predefined threshold. When drift is detected, a new model version $M_{v+1}$ is trained and deployed:
\vspace{0.5mm}
\[
M_{v+1} = \text{Train}(X_t, Y_t).
\]

The pipeline supports scalability by maintaining versioned repositories of models and automating retraining cycles. It reduces manual intervention while enabling responsiveness to traffic variations and anomalies. Inspired by Bhatt et al. \cite{bhatt2024towards}, this approach emphasizes sustainable and self-adaptive workflows, ensuring robustness in dynamic environments.

\textbf{Model Manager.}  
As shown in Figure~\ref{fig:digit_architecture}, the Model Manager organizes and stores models essential for the DT, including \textit{communication}, \textit{predictive}, and \textit{environment} models, but is not limited to them. \textit{Communication models} define protocols and behaviors for data exchange, ensuring seamless interaction between components. \textit{Predictive models}, forecast traffic patterns and congestion based on historical and real-time data. \textit{Environment models} account for external factors like weather and time of day, influencing traffic dynamics. These models are accessible to the Digital Twin Manager, with the \textit{code generator} facilitating the translation of abstract models into executable formats, ensuring compatibility with the \textit{simulator} and \textit{actuator}.

\section{Implementation}

The implementation of the \digit platform follows the architecture outlined in Section~\ref{sec:architecture}, operationalizing its components to enable real-time traffic monitoring, prediction, and visualization for ITS. By leveraging a modular and scalable design, the system ensures integration between data acquisition, predictive analytics, simulation, and user-facing dashboards. 

Real-time traffic data was obtained using IoT-enabled sensors deployed near the IIIT Hyderabad campus, capturing key attributes such as timestamps, flow rates, average speeds, and congestion levels. These sensors, corresponding to the \textit{Sensors on the road} in the architecture (Figure~\ref{fig:digit_architecture}), transmitted preprocessed data to a public API hosted on a Virtual Private Server (VPS). This API acted as the primary data pipeline, ensuring scalability and seamless integration of real-time and historical data. This data was further communicated to the \textit{DT Manager} via the \textit{Communication Layer} as shown in (Figure~\ref{fig:digit_architecture}).

Traffic \textit{Simulator}, a critical component of the \textit{Digital Twin Manager} (Section~\ref{sec:architecture}), was implemented using the SUMO. SUMO utilized the processed sensor data to model transportation networks and simulate real-world scenarios, such as peak traffic hours, roadblocks, and intervention strategies. These simulations allowed for scenario-based testing of adaptive traffic management measures, including dynamic signal timing and rerouting strategies, in alignment with the behavioral models defined in the DCM (Section~\ref{sec:architecture}).

Predictive analytics was powered by ML models, which were handled by the \textit{Model Manager}, as described in Section~\ref{sec:architecture}. Specifically, Long Short-Term Memory (LSTM) and Bidirectional LSTM (BiLSTM) networks, implemented in Python using TensorFlow, were used. These models were designed to analyze temporal patterns in traffic data and provide short-term forecasts, such as predicting congestion levels at specific intersections. The models were integrated with an \textit{MLOps pipeline} to automate processes such as training, validation, and deployment. This pipeline continuously monitored model performance using metrics like Root Mean Squared Error (RMSE). Retraining was dynamically triggered when performance degradation exceeded predefined thresholds, ensuring sustained accuracy under evolving traffic conditions, as described in Section~\ref{sec:architecture}.

The \textit{visualization layer} was implemented using a React-based frontend and a Node.js backend, connected to the API for real-time updates. The dashboard provided an intuitive interface for monitoring key metrics such as vehicle counts, average speeds, and traffic intensity. It also visualized simulation results and predictions through interactive graphs and heatmaps, reflecting the actionable insights derived from the DT's predictive and simulation engines. This implementation extended the DCM’s state and behavioral models into user-facing applications, ensuring operational transparency and stakeholder engagement. The system leveraged the cloud infrastructure for scalable storage and computational resources, supporting both real-time analytics and batch processing. This infrastructure enabled seamless integration across all layers of the architecture, ensuring robust and efficient operation. Overall, the implementation demonstrated the practical realization of the proposed architecture, maintaining synchronization between physical and digital systems while addressing the complex requirements of ITS.

\section{Evaluation}

The evaluation of the architecture of \digit platform is guided by three key aspects: 

\begin{itemize}
    \item \textit{Accuracy}: The predictive performance of the analytics component, measured against observed traffic flow data.  
    \item \textit{Fidelity}: The ability of the simulation environment to replicate real-world traffic patterns and responses to interventions.
    \item \textit{Efficiency}: The computational performance of the system, including the time required to execute predictions and run simulations.  
\end{itemize}

\subsection{Data Collection and Setup}

The data used for evaluating \digit was collected from IoT-enabled sensors deployed at intersections near the IIIT Hyderabad campus, as described in Section~\ref{sec:usecase}. These sensors capture traffic parameters, including vehicle counts, speeds, and congestion levels, at fixed intervals of \textit{5 minutes}. This frequency ensures that the system captures real-time variations in traffic conditions, providing a foundation for both predictive analytics and simulations. The collected data is transmitted via the \textit{Communication Layer} to the \textit{Data Lake}, where it is preprocessed to handle missing values, normalize readings, and structure data for modeling. The dataset spans several days, encompassing both peak and non-peak traffic conditions, making it suitable for evaluating system performance across varying levels of congestion. Using this detailed data set, the DT ensures accurate modeling of traffic dynamics, allowing more realistic simulations and precise predictions. To develop and validate predictive models, the data set was divided into a \textit{ training set (75\%)}, \textit{validation set (15\%)}, and a \textit{testing set (10\%)}. The data was structured into sequences of \textit{15 input timesteps} to predict the \textit{16th timestep}, reflecting the temporal dependencies required for short-term forecasting. For the current implementation, the model takes traffic flow data as input and outputs predicted flow values for future time intervals. 

\subsection{Predictive Model Validation: Accuracy}  

The predictive analytics component of the architecture of \digit platform employs deep learning models, specifically Long Short-Term Memory (LSTM) and Bidirectional Long Short-Term Memory (BiLSTM), implemented within the \textit{MLOps Pipeline}, as shown in Figure~\ref{fig:digit_architecture}. These models are designed to capture temporal dependencies in traffic data, enabling accurate short-term predictions of traffic flow and congestion patterns. Both models were trained using preprocessed data collected from IoT-enabled sensors, as described in Section~\ref{sec:usecase}. The data was structured into sequences of \textit{15 input timesteps} to predict \textit{16th timesteps}, reflecting the temporal dependencies required for short-term forecasting. Traffic flow rates, measured in vehicles per 5 minutes, were used as the target variable to ensure precise predictions suitable for real-time decision-making and intervention strategies. For example, the models can predict increased congestion at an intersection, triggering appropriate interventions such as rerouting traffic or adjusting signal timings. The performance of the LSTM and BiLSTM models was evaluated using standard metrics, including Mean Absolute Error (MAE) (vehicles), Root Mean Squared Error (RMSE) (vehicles), and Mean Absolute Percentage Error (MAPE) (percentage). The results are presented in Table~\ref{tab:model_metrics}.  

\begin{table}[h!]
\centering
\caption{Performance Metrics for Predictive Models}
\label{tab:model_metrics}
\begin{tabular}{|c|c|c|c|}
\hline
\textbf{Model} & \textbf{RMSE (vehicles)} & \textbf{MAE (vehicles)} & \textbf{MAPE (\%)} \\ \hline
LSTM           & 25.522                   & 18.394                  & 13.3               \\ \hline
BiLSTM         & 24.451                   & 17.255                  & 19.1               \\ \hline
\end{tabular}
\end{table}

\begin{figure}[ht!]
    \centering
    \includegraphics[width=\linewidth]{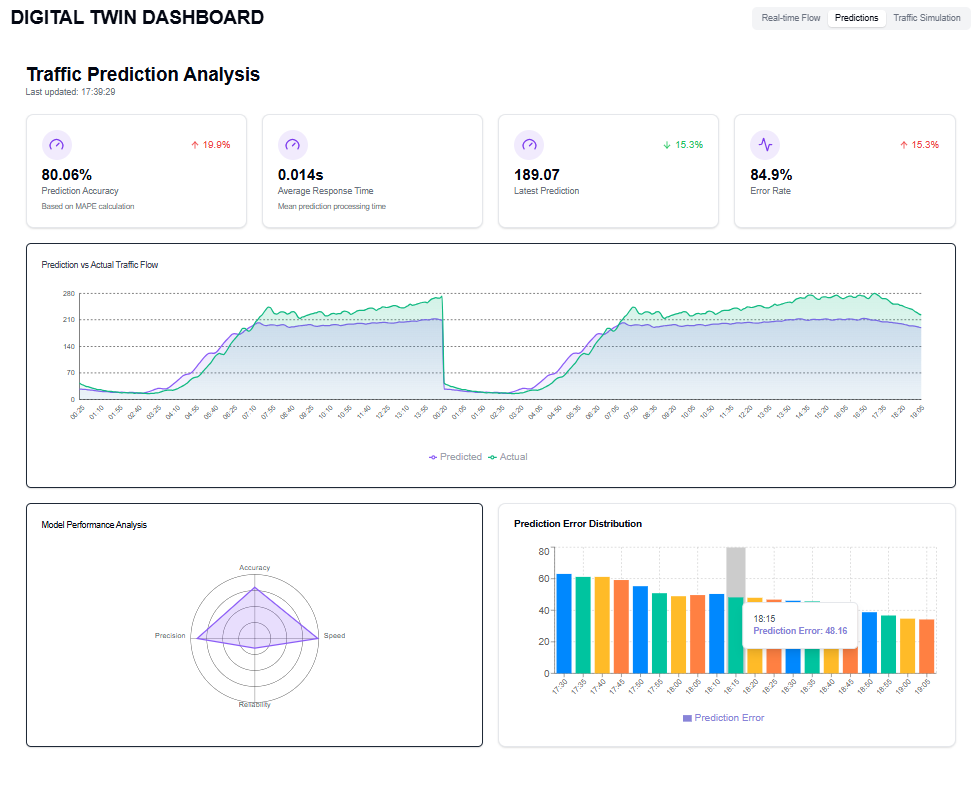}
    \caption{Traffic Prediction Dashboard displaying accuracy, prediction errors, and performance metrics.}
    \label{fig:traffic_prediction_dashboard}
\end{figure}

\noindent The results demonstrate that both models effectively capture traffic dynamics. The BiLSTM model achieved slightly lower RMSE and MAE values, indicating higher predictive accuracy. However, the LSTM model exhibited a lower MAPE, suggesting it may be more robust for percentage-based error evaluation. These findings highlight the complementary strengths of the two models, depending on specific performance criteria. Predictions generated by the models were visualized through the \textit{Visualization Layer} using the \textit{Dashboard for predictions}, as shown in Figure~\ref{fig:traffic_prediction_dashboard}, providing insights into both immediate and future traffic patterns. These insights allow stakeholders to assess the effectiveness of interventions in real-time, ensuring timely and informed traffic management decisions.

\subsection{Simulation Fidelity}  

\begin{figure}[ht!]
    \centering
    \includegraphics[width=\linewidth]{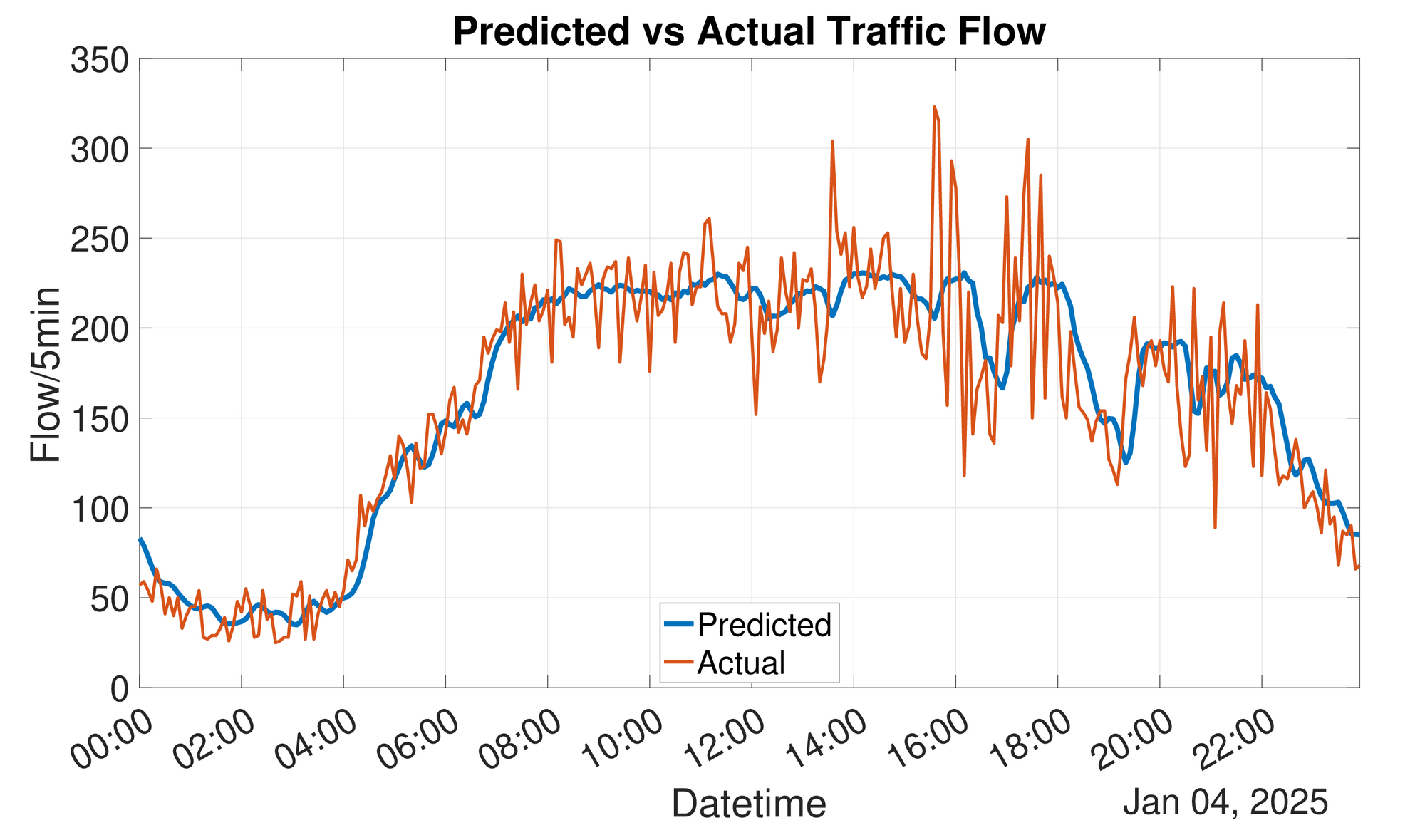}
    \caption{Predicted vs Actual Traffic Flow.}
    \label{fig:pred_vs_actual}
\end{figure}

\begin{figure}[ht!]
    \centering
    \includegraphics[width=\linewidth]{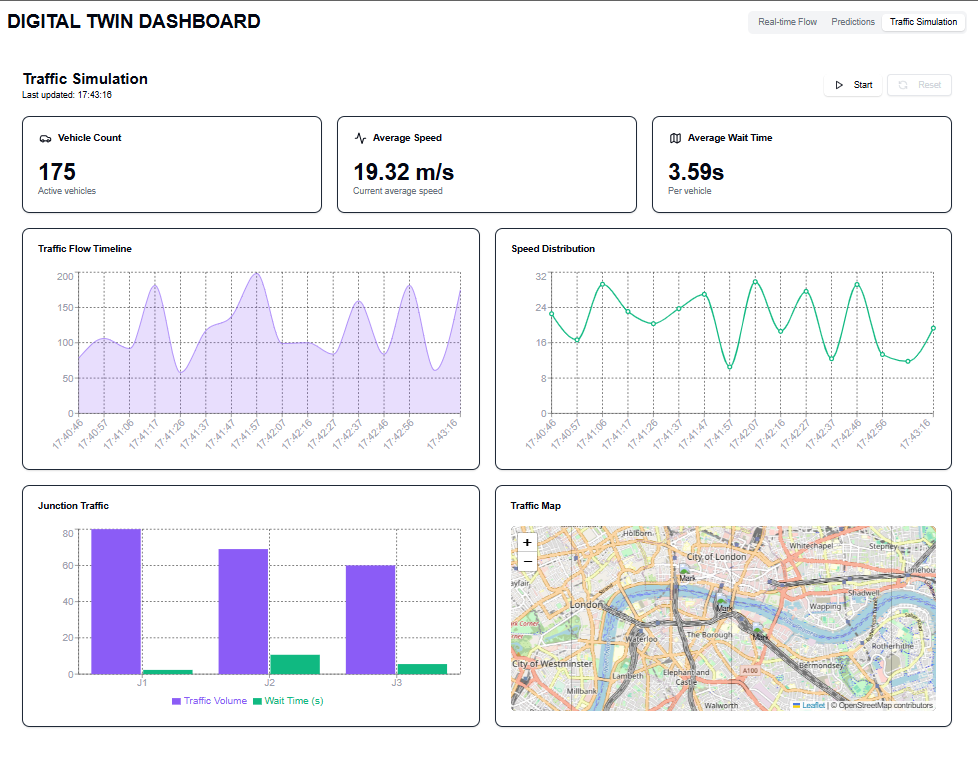}
    \caption{Simulation Dashboard displaying traffic flow timelines, speed distributions, and junction traffic visualizations.}
    \label{fig:simulation_dashboard}
\end{figure}

The simulation fidelity of the architecture of \digit platform was evaluated by analyzing its ability to replicate observed traffic patterns and predict traffic flow dynamics accurately. This evaluation combined machine learning predictions with simulation outputs to validate the system’s ability to model real-world traffic behavior effectively. Traffic predictions were generated using Long Short-Term Memory (LSTM) and Bidirectional LSTM (BiLSTM) models, implemented within the \textit{MLOps Pipeline}. These predictions were compared against actual traffic flow data collected at 5-minute intervals from IoT-enabled sensors, as shown in Figure~\ref{fig:pred_vs_actual}. The results demonstrated a strong correlation between predicted and observed values, capturing key traffic features such as congestion buildup, clearance, and peak-hour variations. To ensure consistency, the modeling assumptions between SUMO simulations and the DT were carefully aligned, including vehicle flow rates, signal timing protocols, and road network configurations. For example, vehicle flow rates from IoT sensors were matched with the input parameters in SUMO to reflect observed traffic volumes. This alignment ensures that the virtual simulations reflect real-world dynamics as closely as possible. Some spikes were observed during periods of high congestion, as shown in Figure~\ref{fig:pred_vs_actual}, due to insufficient data for such scenarios; however, the model is expected to improve with additional data collection. Simulations were executed using the SUMO platform, integrated within the \textit{Simulation and Digital Twin Manager}, to model virtual scenarios reflecting real-world dynamics. These simulation outputs were visualized in the \textit{Visualization Layer} via an interactive dashboard, providing insights into traffic behavior and intervention strategies, as shown in Figure~\ref{fig:simulation_dashboard}.

\subsection{Computational Efficiency}  

The computational performance of \digit was evaluated to validate its suitability for real-time applications. The assessment measured execution times for both predictive modeling and simulation tasks, ensuring responsiveness under time-sensitive conditions. The LSTM and BiLSTM models, implemented within the \textit{MLOps Pipeline}, required an average of \textit{7 milliseconds} to process 15 input timesteps and generate predictions for the next 16 timesteps. Simulations covering the same interval were executed within \textit{15 seconds} using SUMO. These results demonstrate that the system operates within time constraints appropriate for real-time traffic management, enabling timely responses to evolving traffic conditions. The integration of the \textit{MLOps Pipeline} further enhances computational efficiency by automating the retraining and deployment of predictive models. This process ensures that the system adapts dynamically to changes in traffic behavior without manual intervention, maintaining both accuracy and scalability as data volumes increase.


\section{Discussion and Conclusion}  

This paper presented the architecture for the \digit platform. The platform integrates predictive modeling, simulations, and automated workflows to address real-time traffic management challenges. The \textit{DCM} models components such as vehicles, sensors, and communication networks, ensuring seamless data flow between physical and digital systems.  

Evaluation results demonstrated that the platform effectively captured traffic patterns and provided accurate short-term forecasts. Simulations were used to validate the computational efficiency of the platform, showing an average prediction time of 7 ms. The modular design supports scalability, enabling integration of advanced modeling techniques, such as reinforcement learning and hybrid simulations, to enhance adaptability and performance. While the current implementation focuses on traffic forecasting and data-driven modeling, the architecture provides a foundation for expanding capabilities to address broader challenges, including multi-modal transportation systems and communication network modeling. The results validate the feasibility of Digital Twins for traffic management and highlight their potential for improving efficiency and decision-making in dynamic transportation environments.



\section{Future Work}  

\noindent Future efforts will focus on extending the architecture for \digit platform to support multi-modal transportation systems, including public transit and pedestrian flows. Enhancements in modeling communication networks will enable evaluations of key metrics such as latency, throughput, and range, improving performance in connected and autonomous vehicle scenarios. Additional focus will be on integrating adaptive modeling techniques and dynamic calibration methods to enhance scalability during peak traffic and disruptions. Testing across diverse traffic scenarios, including urban roads, highways, and mixed-mode networks, will validate generalizability. Finally, advancements in real-time analytics leveraging IoT and 5G technologies will be explored to strengthen data acquisition, communication reliability, and responsiveness, ensuring the platform can handle dynamic traffic environments effectively.

\section{Acknowledgments}
This research was conducted as part of the UKIERI-SPARC project titled “DigIT—Digital Twins for Integrated Transportation Platform”. The grant number is UKIERI-SPARC/01/23.

\bibliographystyle{plainnat}
\bibliography{references}
\end{document}